\documentclass[final,5p,times,twocolumn,numbers]{elsarticle}

\usepackage{amssymb}
\usepackage{amsmath}

\usepackage{booktabs}
\usepackage{multirow}
\usepackage{hyperref}
\usepackage[justification=centering]{caption}
\usepackage{float}
\usepackage{algorithm}
\usepackage{algpseudocode}

\journal{Computer Methods and Programs in Biomedicine}

\begin{document}
\renewcommand{\figurename}{Fig.}

\begin{frontmatter}



\title{QFedPolyp: A Communication- and Inference-Efficient Federated Learning Framework for Polyp Segmentation} 




\author{Madan Baduwal}
\ead{mb4239@msstate.edu}

\author{Priyanka Paudel}
\ead{pp918@msstate.edu}

\affiliation{
    organization={Department of Computer Science and Engineering},
    institution={Mississippi State University},
    city={Mississippi State},
    state={MS},
    country={USA}
}

\cortext[cor1]{Corresponding author.
Email: mb4239@msstate.edu,
Tel: +1-432-316-1183}

\begin{abstract}

\noindent \textbf{Background and Objective:}
Automatic polyp segmentation supports computer-aided diagnosis and early colorectal cancer detection. Centralized deep learning requires hospitals to share sensitive medical data, while federated learning preserves privacy but introduces high communication costs through repeated transmission of full-precision model parameters. We propose QFedPolyp, a communication- and inference-efficient federated learning framework for collaborative polyp segmentation.

\noindent \textbf{Methods:}
QFedPolyp combines quantization-aware training with low-precision model communication. Each hospital locally trains a lightweight U-Net on private data while simulating quantization during training. Clients transmit quantized model parameters to a central server, where they are reconstructed and aggregated using Federated Averaging. Evaluation is performed on Kvasir-SEG, CVC-ClinicVideoDB, PolypGen, and BKAI-IGH NeoPolyp.

\noindent \textbf{Results:}
Full-precision federated training achieves Dice scores of 0.910 on Kvasir-SEG and 0.930 on CVC-ClinicVideoDB. Uniform 8-bit communication reduces transmission cost by approximately $4\times$ while preserving competitive segmentation accuracy. Quantized models also achieve up to $1.5\times$ faster inference than full-precision models.

\noindent \textbf{Conclusions:}
QFedPolyp enables privacy-preserving collaborative polyp segmentation with reduced communication overhead and faster inference. The resulting lightweight models are suitable for real-time clinical deployment.

\end{abstract}


\begin{keyword}
Federated learning \sep Polyp segmentation \sep Quantization \sep Privacy preservation \sep Communication efficiency \sep U-Net
\end{keyword}

\end{frontmatter}










\section{Introduction}

Colorectal cancer is one of the most common and life-threatening gastrointestinal diseases worldwide. According to the National Cancer Institute, colorectal cancer accounts for a substantial portion of global cancer-related deaths, and early detection plays a critical role in improving patient survival rates \cite{nci2025}. Colonoscopy is the primary clinical procedure used to identify colorectal polyps, which are abnormal tissue growths that may evolve into malignant tumors if left untreated. However, manual inspection of colonoscopy videos is a time-consuming process and is prone to human error, particularly for small or visually subtle lesions. As a result, computer-aided diagnosis (CAD) systems based on deep learning have gained significant attention for assisting clinicians in detecting and segmenting polyps automatically.

Deep convolutional neural networks have demonstrated strong performance in medical image segmentation tasks. Among these architectures, U-Net has become one of the most widely used models due to its encoder–decoder structure and 
\begin{figure}[t]
    \centering
    \vspace{-5pt}
    \includegraphics[width=\linewidth]{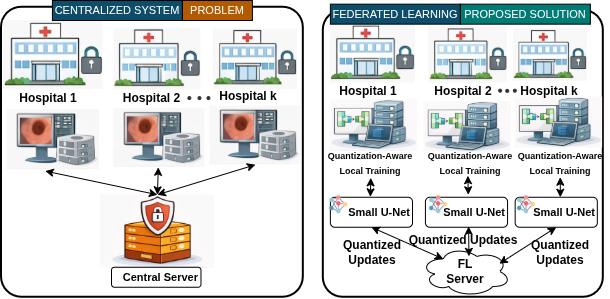}
    \caption{Centralized training versus the proposed QFedPolyp framework. Hospitals perform quantization-aware local training with a lightweight U-Net and send quantized updates to the FL server, preserving privacy while reducing communication cost and enabling efficient inference.}
    \vspace{-8pt}
    \label{fig:problem_solution}
\end{figure}
skip connections that effectively capture both contextual and spatial information \cite{ronneberger2015unet}. Several improved architectures have been proposed specifically for polyp segmentation. For example, PraNet introduces a reverse attention mechanism to refine segmentation boundaries \cite{fan2020pranet}, while HarDNet-MSEG focuses on lightweight model design for efficient real-time segmentation \cite{huang2021hardnetmseg}. These models have achieved strong results on benchmark datasets such as Kvasir-SEG, CVC-ClinicVideoDB, PolypGen, and BKAI-IGH NeoPolyp.

Despite these advances, most deep learning approaches rely on centralized training, where data from multiple hospitals must be collected and stored in a single location. In practical healthcare environments, however, such centralized data aggregation is often infeasible due to strict privacy regulations, institutional policies, and ethical concerns related to patient data sharing. Hospitals are frequently unable or unwilling to transfer sensitive medical images across institutional boundaries, which limits the availability of large multi-institutional datasets for training deep learning models.

Federated learning (FL) has emerged as a promising paradigm for enabling collaborative model training while preserving data privacy. Instead of transferring raw data, federated learning allows multiple institutions to train a shared global model while keeping their data locally stored \cite{mcmahan2017fl}. In this framework, each participating client performs local optimization on its private dataset and sends model updates to a central server. The server aggregates these updates to improve the global model. This decentralized learning paradigm enables collaborative knowledge sharing across hospitals without violating data protection constraints.

Although federated learning effectively addresses privacy concerns associated with centralized training, it introduces a new challenge related to communication efficiency. Federated learning requires repeated exchange of model parameters between participating clients and the central server across multiple communication rounds. Modern deep neural networks often contain millions of parameters, and transmitting full-precision model updates can impose substantial communication overhead, particularly when institutions operate under limited network bandwidth \cite{konevcny2016strategies, chen2021communication}. Consequently, communication cost becomes a critical bottleneck in large-scale federated learning systems.

Several techniques have been proposed to improve communication efficiency in federated learning, including gradient sparsification, model compression, periodic averaging, and parameter quantization \cite{reisizadeh2020fedpaq}. Among these approaches, quantization is particularly effective because it directly reduces the number of bits required to represent model parameters. By transmitting low-precision parameters instead of full-precision floating-point values, the communication cost between clients and servers can be significantly reduced.

However, naive post-training quantization may introduce quantization noise that negatively affects model accuracy. Quantization-aware training (QAT) addresses this limitation by incorporating simulated quantization operations during the training process \cite{jacob2018quantization}. In QAT, the neural network learns parameter distributions that remain robust under low-precision representations, allowing the model to maintain strong performance even when weights are represented using reduced numerical precision.

Figure~\ref{fig:problem_solution} illustrates the conceptual difference between centralized training and the proposed federated learning framework. In centralized systems, hospitals must transfer raw medical images to a central server, creating privacy risks and data transfer challenges. In contrast, the proposed approach enables each hospital to perform quantization-aware local training using a lightweight U-Net model while transmitting only quantized model updates to the federated server. This design preserves patient privacy, reduces communication overhead, and produces efficient models suitable for real-time clinical deployment.

Motivated by these challenges, this paper proposes QFedPolyp, a communication- and inference-efficient federated learning framework for collaborative polyp segmentation. QFedPolyp integrates quantization-aware training with quantized model-parameter transmission to enable bandwidth-efficient federated optimization while maintaining high segmentation accuracy.

The main contributions of this work are summarized as follows:

\begin{itemize}
\item We propose QFedPolyp, a privacy-preserving federated learning framework that enables collaborative polyp segmentation across multiple hospitals without sharing raw medical images.
\item We integrate quantization-aware training into the federated learning pipeline to improve robustness against quantization noise during distributed training.
\item We introduce communication-efficient parameter exchange by transmitting quantized model updates, significantly reducing bandwidth requirements in federated learning.
\item We demonstrate that the resulting low-precision models enable faster inference, making the framework suitable for real-time clinical applications.
\end{itemize}

The implementation of the proposed framework is publicly
available at: \url{https://madanbaduwal.github.io/QFedPolyp}.

The remainder of this paper is organized as follows.
Section~\ref{sec:related_work} reviews related research on deep learning methods for polyp segmentation, federated learning in medical imaging, communication-efficient federated learning, and quantization-aware training.
Section~\ref{sec:methodology} presents QFedPolyp, including its federated optimization strategy, lightweight U-Net architecture, quantization-aware training, and quantized model communication.
Section~\ref{sec:experiments} describes the experimental setup, datasets, evaluation metrics, and performance evaluation, including a detailed comparison with state-of-the-art methods.
Finally, Section~\ref{sec:conclusion} concludes the paper and discusses potential future research directions.

\section{Related Work}
\label{sec:related_work}

Early deep learning approaches have greatly advanced automated polyp segmentation. The U-Net architecture \cite{Ronneberger2015}, first introduced in 2015, became a seminal model for biomedical image segmentation with its encoder-decoder design and skip connections. Building upon U-Net, various improvements have been proposed. For example, UNet++ \cite{Zhou2018} incorporates nested and dense skip pathways to better fuse multi-scale features, yielding more precise segmentations. More recently, researchers have designed specialized CNN architectures targeting colonoscopy polyp segmentation. ResUNet++ \cite{Jha2021} augments the U-Net backbone with residual blocks and applies post-processing (conditional random fields and test-time augmentation) to significantly boost performance on polyp datasets. Fang \emph{et al.} \cite{Fang2019} proposed a Selective Feature Aggregation (SFA) network with dual decoder branches to jointly refine polyp regions and boundaries, enforcing area-boundary consistency for improved edge localization. The PraNet model by Fan \emph{et al.} \cite{Fan2020} introduces a parallel reverse attention mechanism that iteratively highlights polyp regions and boundary cues, achieving state-of-the-art accuracy on multiple benchmark datasets. Another notable method, ACSNet \cite{Zhang2020}, adaptively selects global vs. local context features based on polyp size, enabling the network to handle the large shape variability of polyps. All these advanced deep learning models have markedly improved polyp segmentation accuracy. However, they typically assume training on centralized datasets, which is often infeasible in medical imaging due to privacy and data-sharing restrictions.

Federated learning (FL) has emerged as a solution to train models on decentralized medical data without compromising patient privacy. Sheller \emph{et al.} \cite{Sheller2020} demonstrated one of the first multi-institutional FL implementations for 3D medical image segmentation, showing that a federated model can approach the accuracy of a centrally trained model while each hospital keeps data locally. Subsequent works have highlighted the promise of FL across various medical imaging tasks. Rieke \emph{et al.} \cite{Rieke2020} provided an overview of how FL can enable large-scale collaborations in digital health, emphasizing its potential to overcome data silos and institutional biases in medical datasets. They also discuss challenges such as data heterogeneity (e.g., differing imaging protocols across sites) and the need for robust aggregation techniques. Kaissis \emph{et al.} \cite{Kaissis2020} further surveyed privacy-preserving machine learning in radiology, including FL combined with techniques like secure aggregation and differential privacy to protect sensitive information. These studies underscore that FL is well-suited for medical imaging, where data are plentiful but fragmented. In the context of endoscopy and polyp segmentation, FL can enable models to learn from colonoscopy images across multiple clinics or hospitals without sharing patient data. While FL has been applied to modalities like MRI, CT, and digital pathology, its use in colonoscopy image segmentation remains relatively nascent. Our work addresses this gap by leveraging FL for polyp segmentation, ensuring that knowledge is shared across institutions without exchanging the raw video frames or images.

A critical challenge in federated learning is the communication overhead of exchanging model updates between clients and the server. The standard FedAvg algorithm \cite{McMahan2017} reduces communication frequency by performing multiple local training epochs before averaging models, yet transmitting full precision neural network weights can still be bandwidth-intensive when models are large. To tackle this, numerous communication-efficient FL techniques have been proposed. One approach is quantization of model updates: for instance, TernGrad \cite{Wen2017} compresses gradients to 3-level ternary values, and QSGD \cite{Alistarh2017} stochastically quantizes gradients with provable error bounds, both achieving significant bandwidth savings. Similarly, signSGD \cite{Bernstein2019} sends only the sign of each gradient (1-bit per value), dramatically cutting communication at the cost of additional variance, which can be mitigated by error feedback. Another strategy is sparsification or pruning of updates: instead of sending all gradients, clients transmit only the largest $k$ gradients or those above a threshold \cite{Lin2018}, and accumulate the rest locally. This gradient sparsification can slash the transmitted data size by orders of magnitude (e.g., 100$\times$ compression) while maintaining model performance through techniques like momentum correction \cite{Lin2018}. These communication-efficient methods are often complementary to FL and have been integrated into federated frameworks to enable scaling to many clients or limited-bandwidth settings. In our work, we build on these ideas by employing model quantization to reduce the communication cost per round, which is especially important when deploying federated segmentation to edge devices or across clinics with bandwidth constraints.

Quantization-aware training (QAT) is a technique to ensure that a neural network remains accurate when its weights and activations are reduced to lower precision, typically for efficient inference on edge hardware. Rather than quantizing a trained model as a post-processing step (which can degrade accuracy), QAT simulates low-bit quantization (e.g., 8-bit integers) during training so that the model learns to compensate for quantization errors \cite{Jacob2018}. This approach has been widely used to produce compact, faster models for deployment without significant loss in performance \cite{Jacob2018}. In distributed and federated learning scenarios, QAT can serve a dual purpose: reducing communication by sending quantized model updates, and yielding a final model that is already optimized for low-precision inference on devices. Recent works have started to incorporate QAT into the FL paradigm. Abdelmoniem and Canini \cite{Abdelmoniem2021} introduced an adaptive quantization scheme for FL, where each client trains with a quantized model whose bit-width is tailored to its device capabilities, mitigating heterogeneity in hardware. Ji and Chen \cite{Ji2022} proposed \textit{FedQNN}, an FL framework that trains neural networks with low-bitwidth weights and activations; their results show that aggressive quantization (down to 4--8 bits) can substantially reduce communication and computation load in IoT settings with only minor accuracy loss. Yoon \emph{et al.} \cite{Yoon2022} address the scenario of clients using different precisions: they developed a progressive de-quantization approach that allows the server to aggregate mixed-precision updates and improve a global model despite heterogeneous quantization across clients. Beyond fixed low-precision training, researchers have also explored mixed-precision strategies in FL. Chen and Vikalo \cite{Chen2024} recently presented a CVPR 2024 method to assign different bit-widths to different layers of a model (e.g., more critical layers in higher precision and others in lower precision) during federated training, adapting these assignments to each client’s resource limits. Such quantization-aware techniques are crucial for deploying federated models on resource-constrained devices, as they ensure efficient inference (e.g., real-time polyp segmentation on a mobile endoscope processor) without sacrificing too much accuracy. In our approach, we leverage QAT within federated polyp segmentation to jointly achieve privacy preservation, communication efficiency, and optimized low-precision inference.

\begin{figure*}[t]
    \centering
    \includegraphics[width=\textwidth]{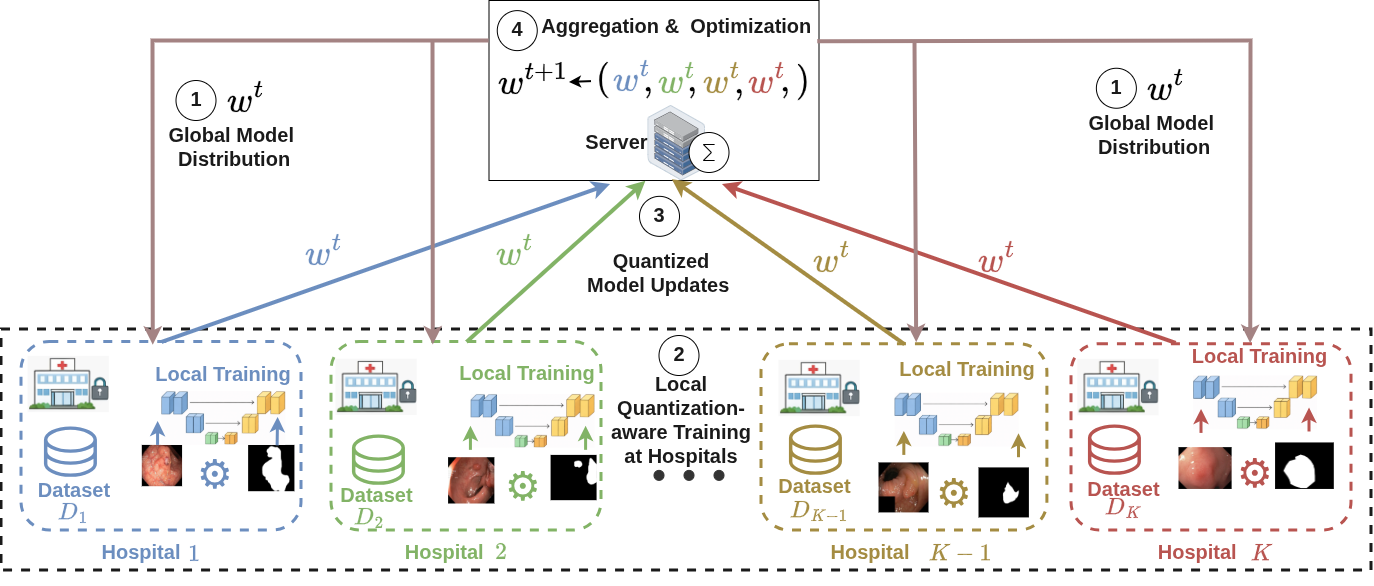}
    \captionsetup{justification=centering}
    \caption{Overview of the proposed QFedPolyp framework for communication- and inference-efficient polyp segmentation. The server distributes the global model to hospitals, each client performs local training on private datasets, transmits quantized model updates, and the server aggregates them using federated averaging to update the global model.}
    \label{fig:fl_architecture}
\end{figure*}

\section{Methods}
\label{sec:methodology}

\subsection{Small U-Net Segmentation Model and Loss Function}

\begin{figure}[H]
    \centering
    \vspace{-5pt}
    \includegraphics[width=\linewidth]{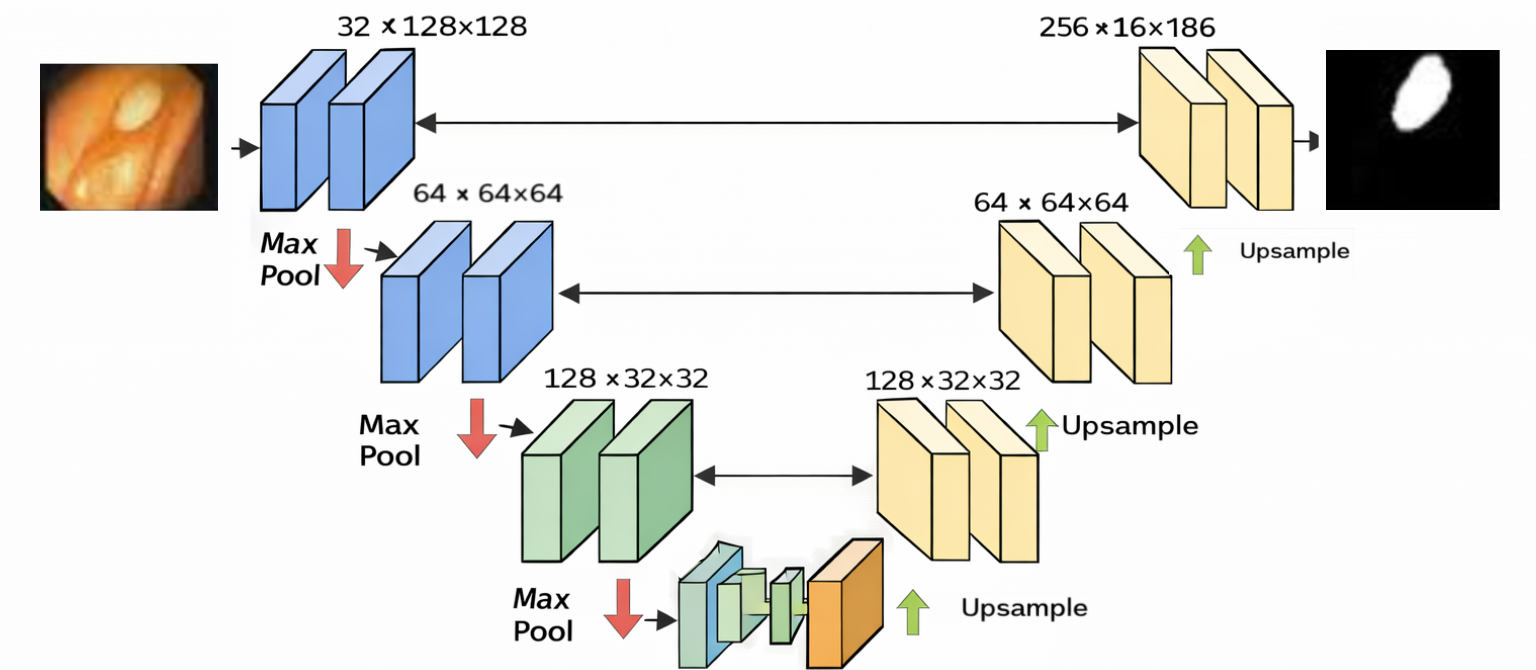}
    \caption{Architecture of the proposed small U-Net used for polyp segmentation. The network follows an encoder-decoder structure with convolutional blocks and max-pooling for feature extraction, followed by upsampling layers and skip connections to recover spatial details. A final $1\times1$ convolution with sigmoid activation produces the binary segmentation mask.}
    \vspace{-8pt}
    \label{fig:small_unet}
\end{figure}

We employ a lightweight U-Net architecture for polyp segmentation, illustrated in Fig.~\ref{fig:small_unet}. The model follows an encoder–decoder structure with symmetric skip connections that enable effective recovery of spatial details while maintaining a compact model size suitable for federated learning environments. Given an input colonoscopy image $x \in \mathbb{R}^{3 \times 128 \times 128}$, the encoder progressively extracts hierarchical feature representations while reducing spatial resolution through max-pooling operations. Each encoder stage consists of two $3\times3$ convolutional layers followed by batch normalization and ReLU activation. The first convolutional block produces feature maps of size $32 \times 128 \times 128$. A $2\times2$ max-pooling operation then reduces the spatial resolution to $64 \times 64$ while increasing the channel dimension to $64$. The next encoder stage further compresses the spatial representation to $128 \times 32 \times 32$ through another max-pooling layer. Finally, a bottleneck layer captures high-level semantic information with feature maps of size $256 \times 16 \times 16$, enabling the network to encode contextual information about the polyp region.

The decoder mirrors the encoder structure and progressively restores spatial resolution through upsampling operations. Starting from the bottleneck representation, feature maps are upsampled by a factor of two and concatenated with the corresponding encoder features through skip connections, as shown in Fig.~\ref{fig:small_unet}. These skip connections help preserve fine-grained spatial information that may otherwise be lost during downsampling. After the first upsampling stage, the decoder produces feature maps of size $128 \times 32 \times 32$, which are fused with encoder features at the same resolution. The reconstruction process continues through subsequent upsampling layers, generating feature maps of size $64 \times 64 \times 64$ and finally $32 \times 128 \times 128$. Each decoder stage includes convolutional refinement layers that combine contextual and spatial information to improve segmentation accuracy.

A final $1\times1$ convolution layer maps the reconstructed feature representation into a single-channel segmentation mask. A sigmoid activation function $\sigma(\cdot)$ converts the network output into per-pixel probabilities indicating the likelihood of each pixel belonging to the polyp region. Formally, for an input image $x$, the predicted segmentation mask is defined as
\begin{equation}
\hat{y} = \sigma(f(x; w)),
\end{equation}
where $f(x;w)$ denotes the output logits of the U-Net model parameterized by weights $w$, and $\hat{y}(i,j) \in [0,1]$ represents the predicted probability that pixel $(i,j)$ belongs to the polyp class.

To train the segmentation network, we employ the Dice loss, which directly optimizes the overlap between predicted and ground-truth segmentation masks. Let $y$ denote the ground-truth mask and $\hat{y}$ the predicted probability map. The Dice loss is defined as
\begin{equation}
\ell_{\text{Dice}}(y,\hat{y}) =
1 -
\frac{2\sum_i y_i \hat{y}_i + \epsilon}
{\sum_i y_i + \sum_i \hat{y}_i + \epsilon},
\end{equation}
where the summation is performed over all pixels and $\epsilon$ is a small smoothing constant to ensure numerical stability. Minimizing the Dice loss encourages the model to maximize spatial overlap between predicted and ground-truth polyp regions, which is particularly suitable for medical image segmentation tasks where foreground regions may occupy a relatively small portion of the image.

\subsection{Federated Optimization using \textit{FedAvg}}

To collaboratively train the lightweight U-Net segmentation model described in the previous section, we adopt a federated learning framework based on the Federated Averaging (FedAvg) algorithm \cite{mcmahan2017fedavg}. The overall training workflow is illustrated in Fig.~\ref{fig:fl_architecture}. In this setting, multiple hospitals jointly train a global segmentation model without sharing raw colonoscopy images or segmentation masks, thereby preserving data privacy.

Assume there are $K$ participating hospitals (clients). Each hospital $k \in \{1,\dots,K\}$ possesses a private dataset
\begin{equation}
D_k = \{(x_i,y_i)\}_{i=1}^{n_k},
\end{equation}
where $x_i$ denotes a colonoscopy image and $y_i$ its corresponding segmentation mask. Let $N = \sum_{k=1}^{K} n_k$ denote the total number of samples across all clients. The objective of federated learning is to optimize the parameters $w$ of the small U-Net model $f(x;w)$ by minimizing the following global objective:

\begin{equation}
\min_{w} F(w) = \sum_{k=1}^{K} \frac{n_k}{N} F_k(w),
\end{equation}

where the local objective for client $k$ is defined as

\begin{equation}
F_k(w) =
\frac{1}{n_k}\sum_{i=1}^{n_k}
\ell\big(f(x_i;w),y_i\big),
\end{equation}

and $\ell(\cdot)$ denotes the Dice loss defined in the previous section. The weighting factor $\frac{n_k}{N}$ ensures that each hospital contributes proportionally to the global model according to its dataset size.

Federated training proceeds through multiple communication rounds between a central server and the participating hospitals. At the beginning of round $t$, the server maintains the current global model parameters $w^{t}$ and distributes them to all clients, as shown in Step~1 of Fig.~\ref{fig:fl_architecture}. Each hospital initializes its local model using the received global parameters:

\begin{equation}
w_k^{t,0} = w^t.
\end{equation}

Subsequently, each client performs local training using its private dataset. During this stage, the small U-Net model is optimized for $E$ local epochs using stochastic gradient descent. The iterative update during local optimization can be written as

\begin{equation}
w_k^{t,j} =
w_k^{t,j-1}
-
\eta \nabla F_k\!\left(w_k^{t,j-1}\right),
\qquad j = 1,\dots,E,
\end{equation}

where $\eta$ denotes the learning rate. As illustrated in Step~2 of Fig.~\ref{fig:fl_architecture}, this local training phase incorporates quantization-aware training (QAT), which simulates low-precision arithmetic during forward and backward passes in order to prepare the model for quantized communication and efficient inference.

After completing local training, each hospital obtains an updated model

\begin{equation}
w_k^{t+1} = w_k^{t,E}.
\end{equation}

Instead of transmitting full-precision parameters, the client compresses its model updates using quantization before sending them to the server. These quantized model updates significantly reduce communication overhead between hospitals and the central server, as shown in Step~3 of Fig.~\ref{fig:fl_architecture}. The quantization strategy used for communication-efficient updates is described in the next subsection.

Upon receiving the client updates, the server reconstructs them and performs aggregation to produce the next global model. Specifically, the updated model parameters are obtained through weighted averaging:

\begin{equation}
w^{t+1} =
\sum_{k=1}^{K}
\frac{n_k}{N}
w_k^{t+1}.
\end{equation}

This aggregation step corresponds to the FedAvg update rule and is illustrated in Step~4 of Fig.~\ref{fig:fl_architecture}. The new global model $w^{t+1}$ is then redistributed to all clients for the next communication round. After $T$ rounds of federated optimization, the model converges toward a solution minimizing the global objective $F(w)$ while ensuring that sensitive medical data remain locally stored at each hospital.

\subsection{Quantization-Aware Training (QAT)}

To enable inference-efficient deployment and improve robustness to low-precision communication, we incorporate quantization-aware training (QAT) into the local optimization process at each participating hospital. As illustrated in Step~2 of Fig.~\ref{fig:fl_architecture}, QAT is performed during the local training phase of the small U-Net segmentation model described in Section~3.1. The key idea is to simulate low-precision arithmetic during training so that the model learns parameters that remain accurate after quantization.

Let $w$ denote the full-precision parameters of the segmentation network $f(x;w)$. During QAT, we introduce a quantization operator $Q_b(\cdot)$ that maps real-valued tensors to a $b$-bit representation. For uniform $b$-bit quantization, the quantized weights $\tilde{w}$ are computed as
\begin{equation}
\tilde{w} = Q_b(w) = \Delta \cdot \mathrm{round}\!\left(\frac{w}{\Delta}\right),
\end{equation}

where $\Delta$ denotes the quantization step size determined by the chosen bit-width $b$. This operation restricts the tensor values to one of $2^b$ discrete levels, effectively simulating integer-based arithmetic used during low-precision inference.

During the forward pass of training, the convolutional weights of the U-Net model are temporarily replaced by their quantized counterparts $\tilde{w}$. In addition, intermediate activations may also be quantized to the same bit precision to emulate realistic inference behavior. Although the model parameters are stored in full precision, this simulated quantization exposes the network to quantization noise during training, allowing it to adapt to reduced numerical precision.

Since the rounding function in the quantization operator is non-differentiable, direct gradient computation is not possible. To address this, we employ the straight-through estimator (STE) during backpropagation. The STE approximates the gradient of the quantization operation by treating it as an identity function, allowing gradients to propagate through the quantization step. Formally, the gradient approximation is written as
\begin{equation}
\frac{\partial L}{\partial w} \approx \frac{\partial L}{\partial \tilde{w}},
\end{equation}

where $L$ denotes the training loss. This approximation enables standard gradient-based optimization while still accounting for quantization effects during the forward pass.

By integrating QAT into the local training process, each hospital learns model parameters that are inherently robust to low-precision representations. Consequently, the final global model can be represented using low-bit integer weights without significant loss in segmentation accuracy. This property is particularly beneficial in the federated learning setting, as it not only enables efficient inference on deployment hardware but also prepares the model for the communication-efficient quantized parameter exchange described in the following subsection.

\subsection{Communication-Efficient Parameter Exchange with $b$-bit Quantization}

A key objective of the proposed framework is to reduce the communication overhead of federated learning while maintaining segmentation accuracy. As illustrated in Step~3 of Fig.~\ref{fig:fl_architecture}, participating hospitals transmit low-precision model parameters to the central server after completing local training. Unlike conventional federated learning approaches that exchange full 32-bit floating-point weights, our framework leverages quantization-aware training (QAT) so that the model parameters are inherently compatible with low-bit representations during training.

Let $w_k^{t+1}$ denote the parameters of the small U-Net model obtained by client $k$ after completing local training in communication round $t$. Because QAT constrains the model parameters to discrete quantization levels during optimization, the resulting weights can be represented directly using $b$ bits. The quantized parameters can be expressed as

\begin{equation}
\tilde{w}_k^{t+1} = Q_b(w_k^{t+1}),
\end{equation}

where $Q_b(\cdot)$ denotes the $b$-bit quantization operator introduced during training and $\tilde{w}_k^{t+1}$ represents the low-precision parameter tensor.

For uniform $b$-bit quantization, the mapping from real-valued weights to discrete levels can be expressed using a linear quantization scheme. Let $w_{\min}$ and $w_{\max}$ denote the minimum and maximum values of a weight tensor. The quantization step size is defined as

\begin{equation}
\Delta = \frac{w_{\max} - w_{\min}}{2^b - 1}.
\end{equation}

Each weight element $w_i$ is mapped to an integer quantization level

\begin{equation}
q_i =
\mathrm{round}
\left(
\frac{w_i - w_{\min}}{\Delta}
\right),
\end{equation}

where $q_i \in \{0,1,\dots,2^b-1\}$ can be stored using $b$ bits. Because these quantization levels are already incorporated during QAT, the client can transmit the low-precision parameters directly to the server without performing an additional compression step.

After receiving the quantized model parameters from all participating hospitals, the server performs the standard FedAvg aggregation to update the global model. Since the model was trained with quantization-aware training, the parameters are naturally robust to low-precision representation, ensuring that segmentation performance is preserved despite the reduced communication precision.

Using $b$-bit parameter exchange significantly reduces the communication cost during federated learning. Compared with the standard 32-bit floating-point representation, the communication requirement per model update decreases by approximately a factor of $32/b$. For example, using $b=8$ leads to roughly a fourfold reduction in transmitted data size while maintaining accurate polyp segmentation. This improvement alleviates the communication bottleneck in federated medical learning and enables efficient collaboration among hospitals with limited network bandwidth.

\begin{algorithm}[t]
\caption{Quantized Federated Learning with FedAvg}
\label{alg:fedavg}
\begin{algorithmic}[1]
\Require $K$ clients indexed by $k$; local datasets $D_k$; communication rounds $T$; initial global model $w^0$; local epochs $E$; learning rate $\eta$.
\Ensure Final global model $w^T$.
\For{$t = 1$ to $T$}
    \For{each client $k = 1$ to $K$ \textbf{in parallel}}
        \State send global model $w^{t-1}$ to client $k$
        \State $w_k^t \leftarrow \text{LocalQAT}(w^{t-1}, D_k, E, \eta)$
        \State choose client-specific bit-width $b_k$
        \State $\tilde{w}_k^t \leftarrow \text{UniformQuantize}(w_k^t, b_k)$
        \Comment{client $k$ applies its own quantization level}
        \State send $\tilde{w}_k^t$ to server
    \EndFor
    \State $N \leftarrow \sum_{j=1}^{K} |D_j|$
    \State $w^t \leftarrow \sum_{k=1}^{K} \frac{|D_k|}{N} \tilde{w}_k^t$
    \Comment{FedAvg aggregation}
\EndFor
\State \Return $w^T$
\end{algorithmic}
\end{algorithm}

\begin{algorithm}[t]
\caption{LocalQAT$(w, D_k, E, \eta)$}
\label{alg:localqat}
\begin{algorithmic}[1]
\Require Initial model $w$; local dataset $D_k$; local epochs $E$; learning rate $\eta$.
\Ensure Updated local model $w$.
\For{$epoch = 1$ to $E$}
    \For{each mini-batch $B \subseteq D_k$}
        \State $\tilde{w} \leftarrow Q(w)$
        \Comment{fake quantization during forward pass}
        \State $L \leftarrow \mathcal{L}(f(B;\tilde{w}))$
        \State $g \leftarrow \nabla_w L$
        \Comment{STE used in backpropagation}
        \State $w \leftarrow w - \eta g$
    \EndFor
\EndFor
\State \Return $w$
\end{algorithmic}
\end{algorithm}

\begin{algorithm}[t]
\caption{UniformQuantize$(x, b_k)$}
\label{alg:quantize}
\begin{algorithmic}[1]
\Require Tensor $x=(x_1,\dots,x_n)$; client-specific bit-width $b_k$.
\Ensure Quantized tensor $\tilde{x}$.
\State $x_{\min} \leftarrow \min_i x_i$
\State $x_{\max} \leftarrow \max_i x_i$
\State $\Delta \leftarrow \frac{x_{\max}-x_{\min}}{2^{b_k}-1}$
\For{$i = 1$ to $n$}
    \State $q_i \leftarrow \mathrm{round}\!\left(\frac{x_i-x_{\min}}{\Delta}\right)$
    \State $\tilde{x}_i \leftarrow x_{\min} + q_i \Delta$
\EndFor
\State \Return $\tilde{x}$
\end{algorithmic}
\end{algorithm}

\section{Results}
\label{sec:experiments}

\subsection{Datasets and Experimental Setup}

We evaluated QFedPolyp on four publicly available colonoscopy
polyp segmentation datasets: Kvasir-SEG (1,000 images), CVC-ClinicVideoDB (11,954 frames), PolypGen (6,282 images), and BKAI-IGH NeoPolyp (1,200 images). Each dataset provides pixel-level ground-truth segmentation masks for polyps and represents colonoscopy data collected under different clinical environments and acquisition protocols. Kvasir-SEG is a widely used benchmark dataset containing diverse polyp appearances with high-quality annotations. CVC-ClinicVideoDB consists of colonoscopy video frames extracted from clinical procedures and provides dense frame-level annotations suitable for evaluating segmentation models under real clinical imaging conditions. PolypGen is a large multi-center dataset collected from several hospitals, offering substantial variability in polyp morphology, imaging devices, and acquisition settings. BKAI-IGH NeoPolyp further introduces additional variability in polyp shapes, sizes, and imaging conditions, making it useful for evaluating the robustness and generalization capability of segmentation models. The diversity of these datasets enables comprehensive evaluation of the proposed method across different clinical scenarios and imaging distributions.

To simulate a realistic multi-hospital federated learning scenario, each dataset was partitioned into $K=5$ disjoint subsets representing independent clients (hospitals). Each client therefore holds a private local dataset $D_k$ that is never shared with other clients or the central server. During training, only model parameters are exchanged between the server and clients, ensuring that raw medical images and annotations remain local to each institution and preserving patient privacy.

All federated experiments were conducted using the standard Federated Averaging (FedAvg) algorithm to aggregate client models on the central server. As the segmentation backbone, we employed a lightweight U-Net architecture described in Section~3.1. The model was intentionally designed to be compact in order to reduce communication overhead and enable efficient deployment on resource-constrained devices while still capturing the multi-scale contextual information required for accurate polyp segmentation. During each communication round, the global model was distributed to all participating clients. Each client then trained its local model using its private dataset for a fixed number of epochs per round (five local epochs in our experiments). Training was performed using the Adam optimizer with a learning rate of $10^{-3}$. The total number of communication rounds was set to $T=150$, which was sufficient for convergence across all experimental configurations. To ensure reproducibility and fair comparison across experiments, a consistent training pipeline and hyperparameters were used, all clients participated in every round, and a fixed random seed of 42 was used.

A key aspect of our framework is the integration of quantization-aware training (QAT) at the client level to enable communication-efficient and inference-efficient model training. During local training, model weights are pseudo-quantized in the forward pass to simulate low-precision computation while gradients are computed using the straight-through estimator. As a result, the trained models become robust to low-precision representation. After local training, each client transmits its quantized model parameters to the server rather than full-precision weights, which significantly reduces communication overhead. In our experiments, uniform quantization was used at the client side to represent model parameters with reduced precision. While an 8-bit configuration was used as the default setting due to its strong balance between accuracy and communication efficiency, our framework also allows different clients to employ different quantization bit-widths depending on their computational or communication constraints. This heterogeneous quantization capability enables more flexible deployment across hospitals with varying hardware capabilities. Compared with standard 32-bit floating-point parameter exchange, transmitting $b$-bit parameters reduces communication cost by approximately a factor of $32/b$. For example, using $b=8$ results in roughly a fourfold reduction in communication payload while maintaining segmentation accuracy.

All training experiments were conducted on a workstation equipped with an NVIDIA RTX 4090 GPU (24\,GB) for acceleration. For inference benchmarking, we evaluated the trained models on both GPU and CPU platforms, including an Intel i9-14900KF processor, in order to assess deployment efficiency across different hardware environments.

\subsection{Evaluation Metrics}

Model performance was evaluated using two widely adopted segmentation metrics: the Dice Similarity Coefficient (DSC) and the Intersection over Union (IoU). These metrics measure the overlap between the predicted segmentation mask and the ground-truth annotation and are commonly used in medical image segmentation tasks.

The Dice Similarity Coefficient measures the similarity between the predicted mask $\hat{Y}$ and the ground-truth mask $Y$. It is defined as

\begin{equation}
\text{DSC} = \frac{2TP}{2TP + FP + FN},
\end{equation}

where $TP$, $FP$, and $FN$ denote the number of true positive, false positive, and false negative pixels, respectively. The Dice score ranges from 0 to 1, where a higher value indicates better overlap between the predicted and ground-truth segmentation.

The Intersection over Union (IoU), also known as the Jaccard Index, measures the ratio between the intersection and the union of the predicted and ground-truth masks. It is defined as

\begin{equation}
\text{IoU} = \frac{TP}{TP + FP + FN}.
\end{equation}

IoU also ranges from 0 to 1, with higher values indicating better segmentation performance.

For each dataset, the metrics were computed on the test images by comparing the predicted segmentation masks with their corresponding ground-truth masks. The final performance values were obtained by averaging the metric scores across all test images.
\subsection{Baseline Comparisons}
We compare the following training strategies:

\noindent$\bullet$\textbf{Centralized training (upper bound):} a conventional scenario where all data from the 5 partitions is pooled and the U-Net is trained as a single model (violating privacy, but gives a benchmark for maximum achievable accuracy).

\noindent$\bullet$\textbf{Vanilla Federated Learning (FL, 32-bit):} our federated setup using full-precision model exchange (no quantization) and no QAT. This represents the baseline federated performance without any communication compression.

\noindent$\bullet$\textbf{QFedPolyp (Proposed):} the proposed framework that integrates quantization-aware training (QAT) with low-precision model communication in the federated learning pipeline.

\noindent$\bullet$\textbf{Ablation -- Uniform Quantization:} a configuration where all participating hospitals transmit model updates using the same bit-width (e.g., 8-bit) to evaluate the performance–communication trade-off under a uniform precision setting.

\noindent$\bullet$\textbf{Ablation -- Hospital-wise Mixed Precision:} a configuration where different hospitals communicate model updates using different bit-widths, simulating heterogeneous clinical environments where institutions may have varying network bandwidth or hardware capabilities.

\subsection{Centralized training}

\begin{figure}[htbp]
    \centering
    \includegraphics[width=0.4\textwidth]{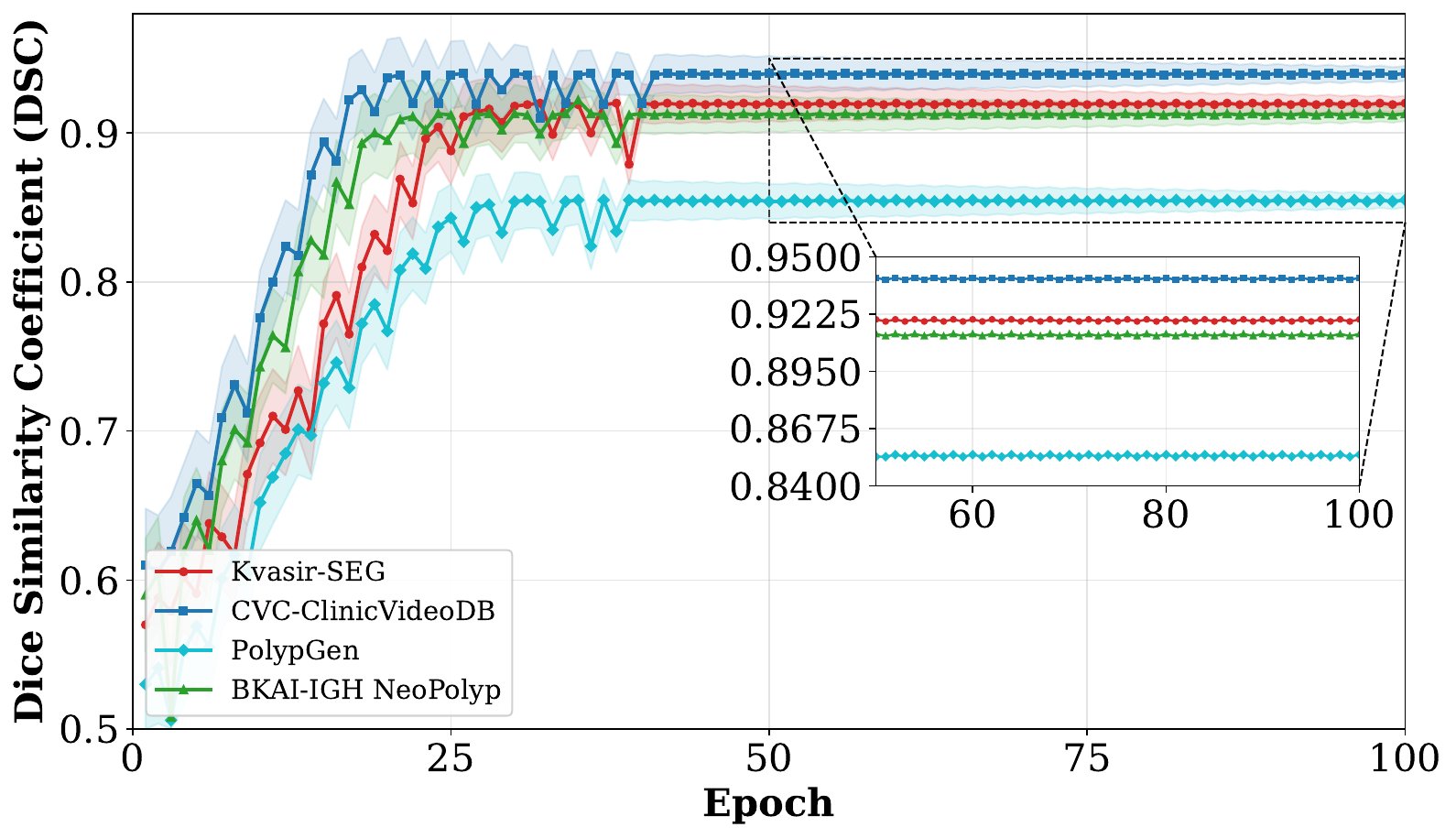}
    \vspace{-5pt}
    \caption{Dice similarity coefficient across the four evaluation datasets under centralized training.}
    \vspace{-5pt}
    \label{fig:dsc_centralized}
\end{figure}

\begin{figure}[htbp]
    \centering
    \includegraphics[width=0.4\textwidth]{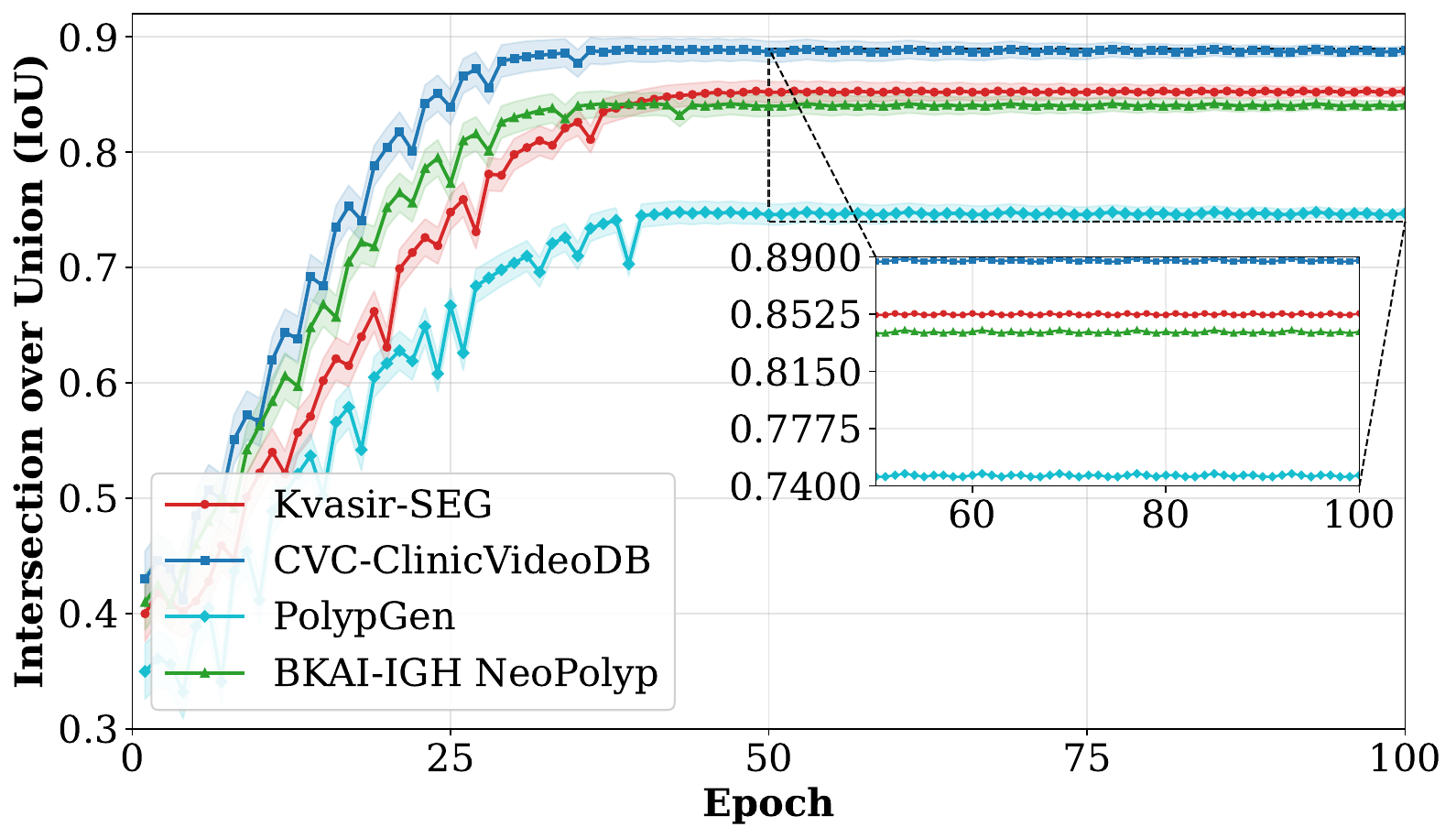}
    \vspace{-5pt}
    \caption{Intersection over Union across the four evaluation datasets under centralized training.}
    \vspace{-5pt}
    \label{fig:iou_centralized}
\end{figure}

Figure~\ref{fig:dsc_centralized} and Figure~\ref{fig:iou_centralized} illustrate the convergence behavior of the segmentation model under centralized training across the four datasets. As observed in the training curves, the model improves rapidly during the early epochs and reaches stable convergence after approximately 30--40 epochs. After this point, the curves flatten and only minor fluctuations are observed, indicating stable optimization and consistent segmentation performance. Among the evaluated datasets, \textit{CVC-ClinicVideoDB} achieves the highest segmentation accuracy, reaching approximately $0.94$ Dice and $0.89$ IoU at convergence. \textit{Kvasir-SEG} and \textit{BKAI-IGH NeoPolyp} demonstrate similar performance levels, achieving Dice scores around $0.92$ and IoU values around $0.84$--$0.85$. In contrast, \textit{PolypGen} remains comparatively more challenging, converging to approximately $0.85$ Dice and $0.75$ IoU, which reflects the higher variability and domain differences present in this dataset.

Overall, the centralized training setup provides the strongest performance across all datasets and serves as an upper-bound reference for the federated and quantized federated learning experiments. The stable plateau observed after convergence further confirms the robustness of the training process and the generalization capability of the segmentation model across different clinical datasets.

\subsection{Full-precision (32-bit) federated performance}

\begin{figure}[htbp]
    \centering
    \includegraphics[width=0.4\textwidth]{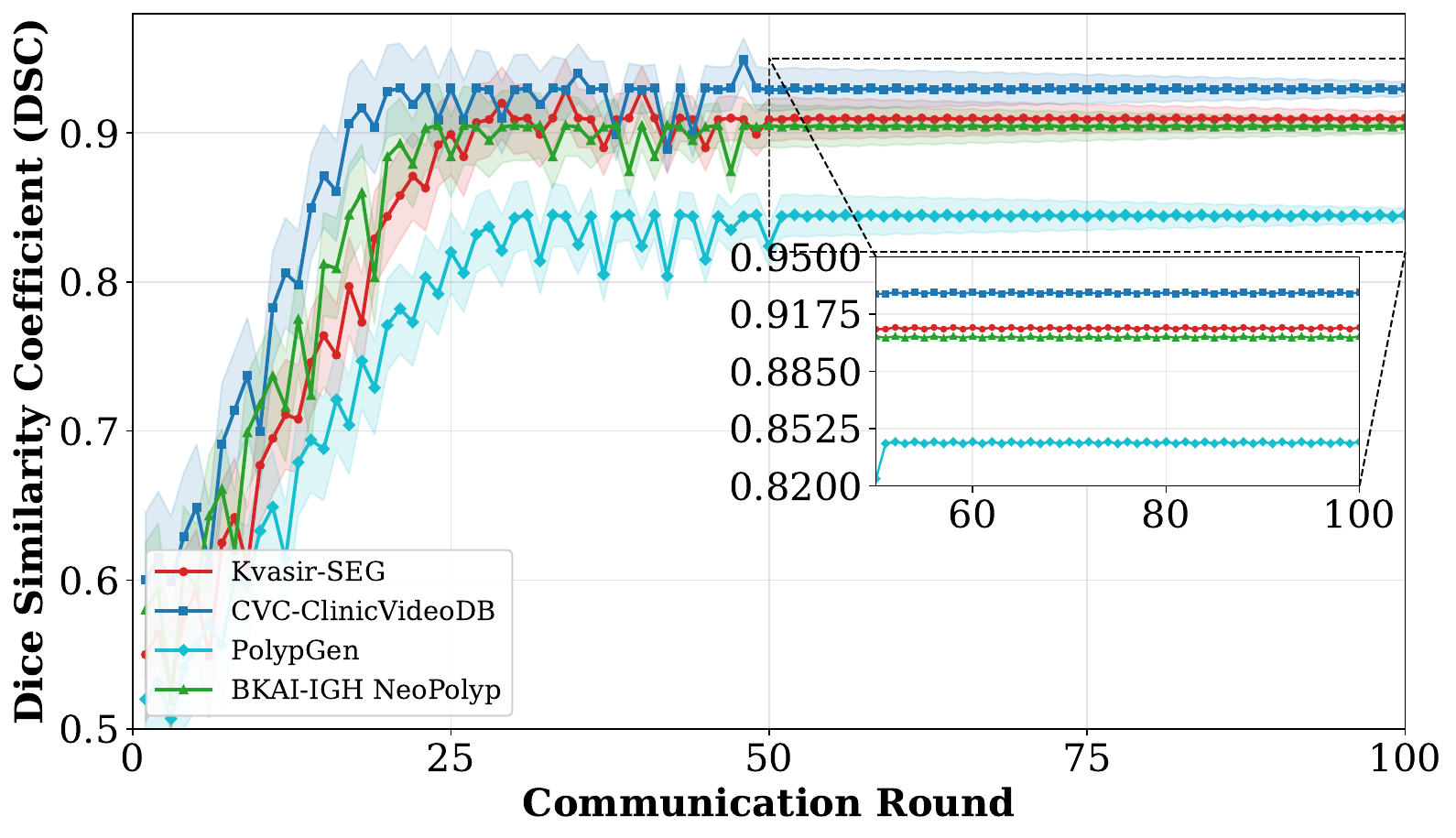}
    \vspace{-5pt}
    \caption{Dice similarity coefficient across the four evaluation datasets under uniform 32-bit federated communication.}
    \vspace{-5pt}
    \label{fig:dsc_uniform32}
\end{figure}

\begin{figure}[htbp]
    \centering
    \includegraphics[width=0.4\textwidth]{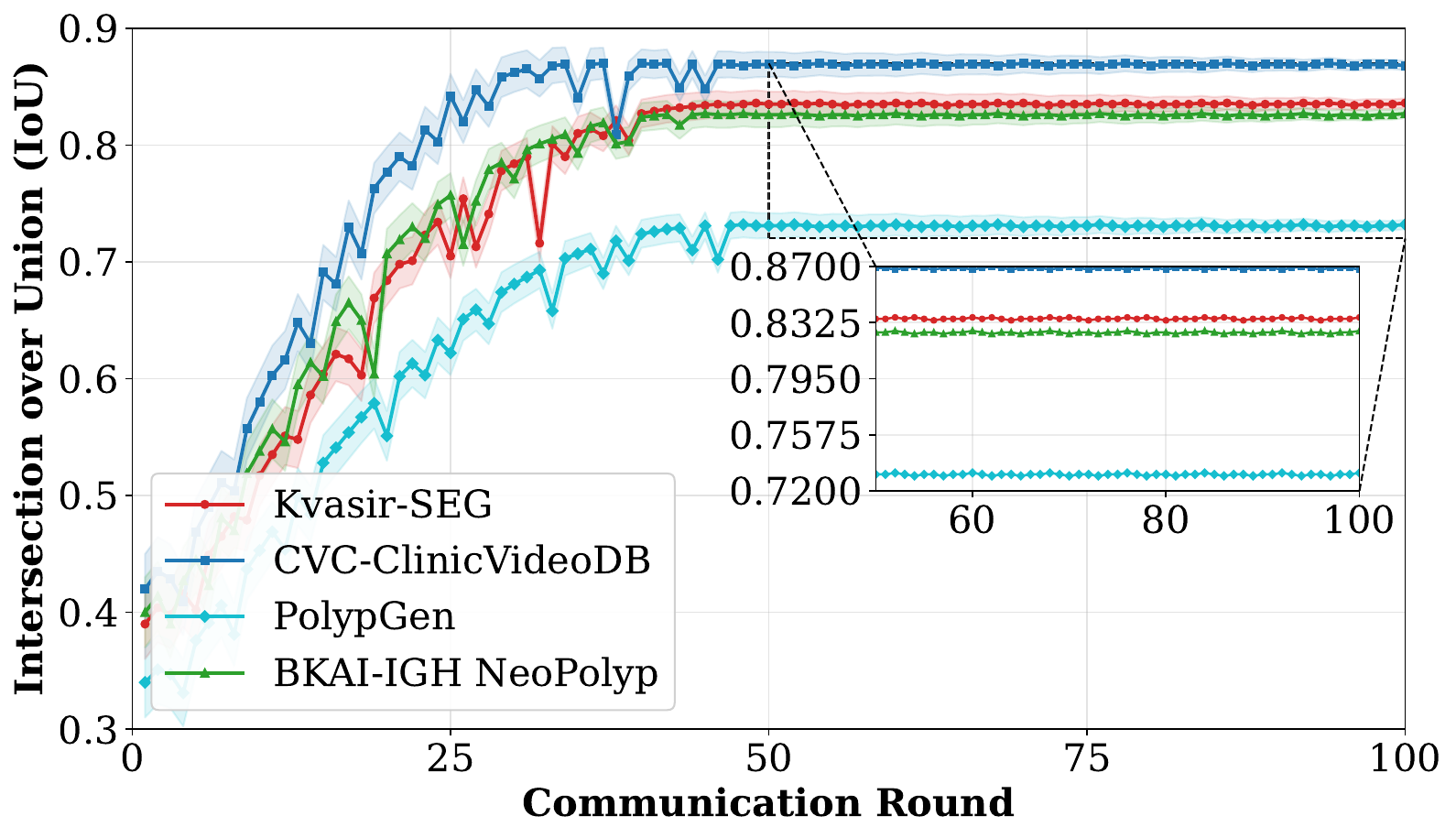}
    \vspace{-5pt}
    \caption{Intersection over Union across the four evaluation datasets under uniform 32-bit federated communication.}
    \vspace{-5pt}
    \label{fig:iou_uniform32}
\end{figure}

Figures~\ref{fig:dsc_uniform32} and~\ref{fig:iou_uniform32} present the segmentation performance under full-precision federated learning where model parameters are exchanged in 32-bit floating-point representation across communication rounds. The global model exhibits a stable convergence pattern, with rapid improvement during the early training rounds followed by gradual stabilization after approximately 50–60 communication rounds. After convergence, the performance curves flatten with only minor fluctuations, indicating consistent collaborative learning among participating clients. Among the evaluated datasets, \textit{CVC-ClinicVideoDB} achieves the highest segmentation performance, converging to approximately $0.93$ Dice and about $0.87$ IoU. \textit{Kvasir-SEG} and \textit{BKAI-IGH NeoPolyp} show comparable results, reaching Dice scores around $0.91$ with IoU values close to $0.83$–$0.84$. In comparison, \textit{PolypGen} remains relatively more challenging, achieving approximately $0.84$ Dice and around $0.73$ IoU, which can be attributed to greater variability in polyp appearance and domain differences across imaging conditions.

Overall, these results demonstrate that full-precision federated learning can achieve segmentation performance close to centralized training while maintaining data privacy by keeping clinical data localized at each institution.

\begin{table*}[!t]
\centering
\caption{Federated learning performance under hospital-wise quantization configurations. Each configuration indicates the bit-width used by the five hospitals. For each dataset, we report Dice Similarity Coefficient (DSC), Intersection over Union (IoU), and the communication round at which the global model converged.}
\label{tab:mixed_quantization_results}
\resizebox{\textwidth}{!}{%
\begin{tabular}{c c c ccc ccc ccc ccc}
\toprule
\multirow{2}{*}{\textbf{Hospitals Quantization}} &
\multirow{2}{*}{\textbf{Avg. Bits}} &
\multirow{2}{*}{\textbf{Comm. Reduction}} &
\multicolumn{3}{c}{\textbf{Kvasir-SEG}} &
\multicolumn{3}{c}{\textbf{CVC-ClinicVideoDB}} &
\multicolumn{3}{c}{\textbf{PolypGen}} &
\multicolumn{3}{c}{\textbf{BKAI-IGH NeoPolyp}} \\
\cmidrule(lr){4-6} \cmidrule(lr){7-9} \cmidrule(lr){10-12} \cmidrule(lr){13-15}
& & & \textbf{DSC} & \textbf{IoU} & \textbf{Round}
      & \textbf{DSC} & \textbf{IoU} & \textbf{Round}
      & \textbf{DSC} & \textbf{IoU} & \textbf{Round}
      & \textbf{DSC} & \textbf{IoU} & \textbf{Round} \\
\midrule

$(4,4,4,4,4)$     & 4.0  & $8\times$  & 0.760 & 0.62 & 104 & 0.780 & 0.64 & 102 & 0.700 & 0.54 & 106 & 0.750 & 0.60 & 104 \\
$(4,4,4,4,8)$     & 4.8  & $6.7\times$& 0.770 & 0.63 & 102 & 0.790 & 0.65 & 100 & 0.710 & 0.55 & 104 & 0.760 & 0.61 & 102 \\
$(4,4,4,8,8)$     & 5.6  & $5.7\times$& 0.780 & 0.64 & 100 & 0.800 & 0.67 & 98 & 0.720 & 0.56 & 102 & 0.770 & 0.62 & 100 \\
$(4,4,8,8,8)$     & 6.4  & $5\times$  & 0.790 & 0.65 & 98 & 0.810 & 0.68 & 96 & 0.730 & 0.57 & 100 & 0.780 & 0.63 & 98 \\
$(4,8,8,8,8)$     & 7.2  & $4.4\times$& 0.799 & 0.67 & 96 & 0.820 & 0.70 & 94 & 0.740 & 0.58 & 98 & 0.790 & 0.65 & 96 \\

$(8,8,8,8,8)$     & 8.0  & $4\times$  & \textbf{0.895} & \textbf{0.81} & 82 
& \textbf{0.915} & \textbf{0.84} & 80 
& \textbf{0.825} & \textbf{0.70} & 86 
& \textbf{0.885} & \textbf{0.80} & 82 \\

$(8,8,8,8,16)$    & 9.6  & $3.3\times$& 0.897 & 0.81 & 80 & 0.918 & 0.85 & 78 & 0.828 & 0.71 & 84 & 0.892 & 0.81 & 80 \\
$(8,8,8,16,16)$   & 11.2 & $2.9\times$& 0.899 & 0.82 & 78 & 0.920 & 0.86 & 76 & 0.830 & 0.72 & 82 & 0.894 & 0.82 & 78 \\
$(8,8,16,16,16)$  & 12.8 & $2.5\times$& 0.901 & 0.82 & 76 & 0.922 & 0.86 & 74 & 0.832 & 0.72 & 80 & 0.896 & 0.82 & 76 \\
$(8,16,16,16,16)$ & 14.4 & $2.2\times$& 0.903 & 0.83 & 74 & 0.924 & 0.86 & 72 & 0.834 & 0.72 & 78 & 0.898 & 0.82 & 74 \\

$(16,16,16,16,16)$& 16.0 & $2\times$  & 0.905 & 0.83 & 72 & 0.926 & 0.87 & 70 & 0.836 & 0.72 & 76 & 0.900 & 0.82 & 72 \\
$(16,16,16,16,32)$& 19.2 & $1.7\times$& 0.907 & 0.83 & 70 & 0.928 & 0.87 & 68 & 0.838 & 0.72 & 74 & 0.902 & 0.82 & 70 \\
$(16,16,16,32,32)$& 22.4 & $1.4\times$& 0.908 & 0.83 & 68 & 0.929 & 0.87 & 66 & 0.840 & 0.72 & 72 & 0.903 & 0.83 & 68 \\
$(16,16,32,32,32)$& 25.6 & $1.25\times$& 0.909 & 0.83 & 66 & 0.9295 & 0.87 & 64 & 0.842 & 0.73 & 70 & 0.904 & 0.83 & 66 \\
$(16,32,32,32,32)$& 28.8 & $1.1\times$& 0.9095 & 0.84 & 64 & 0.9298 & 0.87 & 62 & 0.843 & 0.73 & 68 & 0.9045 & 0.83 & 64 \\

$(32,32,32,32,32)$& 32.0 & $1\times$  & \textbf{0.910} & \textbf{0.84} & \textbf{60} 
& \textbf{0.930} & \textbf{0.87} & \textbf{58} 
& \textbf{0.845} & \textbf{0.73} & \textbf{62} 
& \textbf{0.905} & \textbf{0.83} & \textbf{60} \\

\bottomrule
\end{tabular}%
}
\end{table*}

\subsection{Quantization impact on communication and accuracy}


\begin{figure}[htbp]
    \centering
    \includegraphics[width=0.4\textwidth]{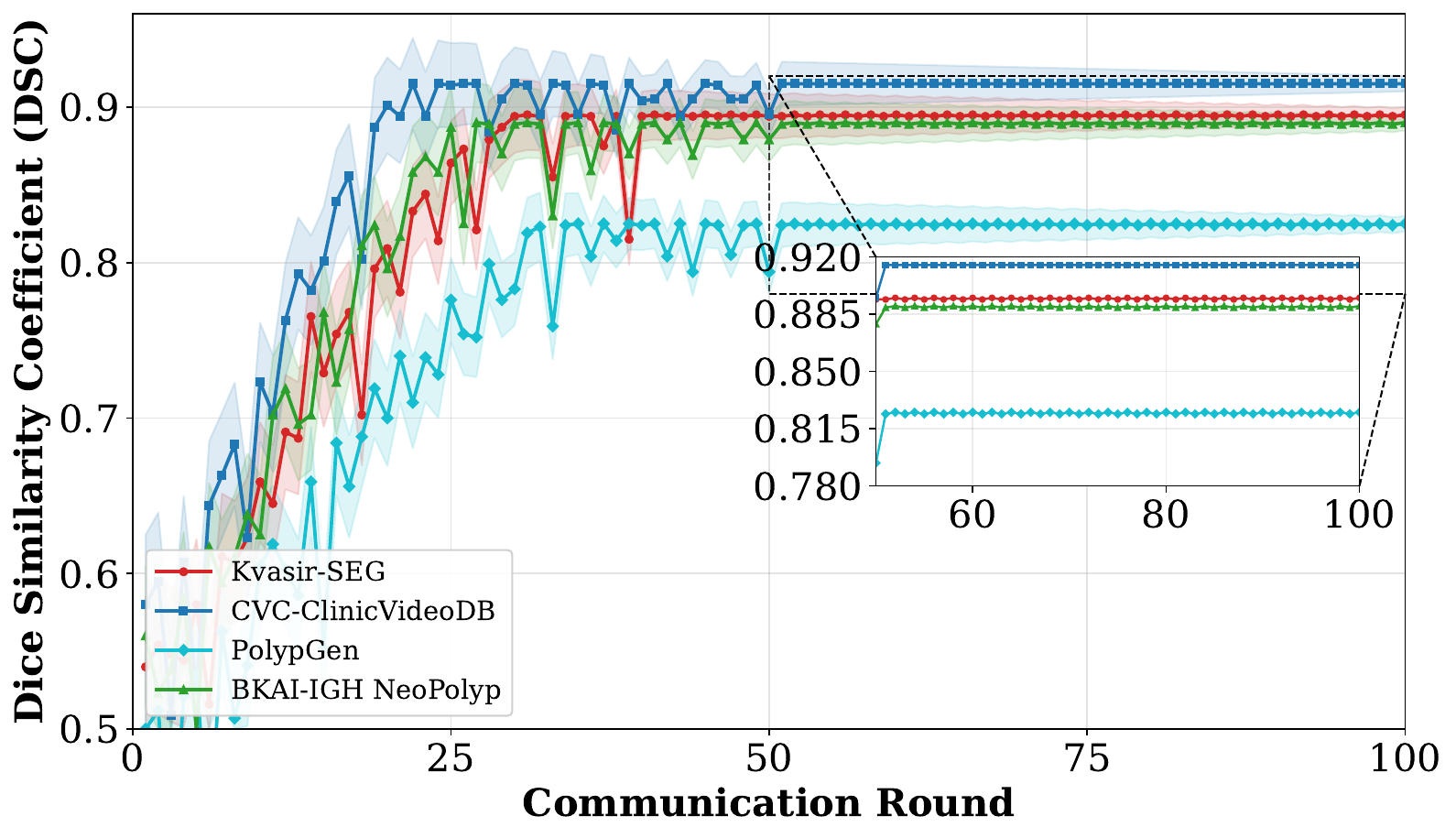}
    \vspace{-5pt}
    \caption{Dice similarity coefficient across the four evaluation datasets under uniform 8-bit federated communication.}
    \vspace{-5pt}
    \label{fig:dsc_uniform8}
\end{figure}

\begin{figure}[htbp]
    \centering
    \includegraphics[width=0.4\textwidth]{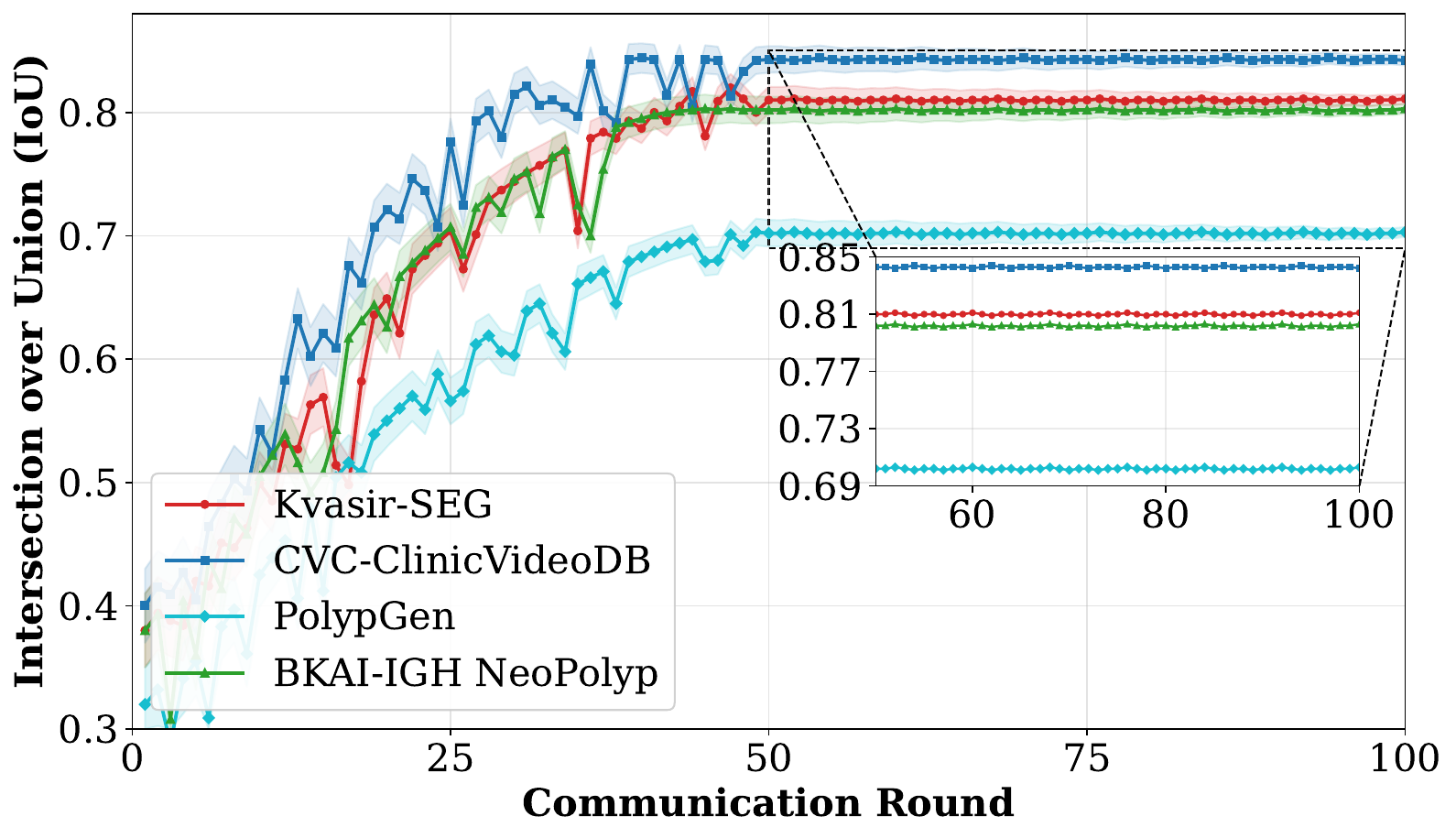}
    \vspace{-5pt}
    \caption{Intersection over Union across the four evaluation datasets under uniform 8-bit federated communication.}
    \vspace{-5pt}
    \label{fig:iou_uniform8}
\end{figure}

Figures~\ref{fig:dsc_uniform8} and~\ref{fig:iou_uniform8} present the segmentation performance of QFedPolyp under uniform 8-bit federated communication. Compared with the full-precision 32-bit federated setting, transmitting quantized parameters significantly reduces the communication payload (approximately $4\times$ smaller) while maintaining competitive segmentation accuracy. From the convergence curves, the model continues to show stable learning behavior across communication rounds, although convergence occurs slightly later compared with the full-precision setting. The global model typically stabilizes after approximately 70--80 communication rounds, after which the performance curves flatten with minimal oscillation. Among the evaluated datasets, \textit{CVC-ClinicVideoDB} again achieves the highest performance, converging to approximately $0.92$ Dice and about $0.84$ IoU. \textit{Kvasir-SEG} reaches around $0.895$ Dice with an IoU close to $0.81$, while \textit{BKAI-IGH NeoPolyp} achieves roughly $0.885$ Dice and about $0.80$ IoU. In comparison, \textit{PolypGen} remains the most challenging dataset, reaching approximately $0.825$ Dice and around $0.70$ IoU due to higher variability and domain differences across imaging environments.

Despite the reduced numerical precision, the degradation in segmentation accuracy remains relatively small compared with the 32-bit federated setting. These results indicate that quantization-aware training effectively compensates for quantization noise during distributed optimization, enabling efficient low-precision communication while preserving strong segmentation performance across heterogeneous clinical datasets.

\subsection{Hospital-wise Mixed-Precision Quantization Analysis}

Table~\ref{tab:mixed_quantization_results} presents a comprehensive analysis of both uniform and mixed-precision quantization strategies across participating hospitals. Each configuration specifies the communication bit-width used by the five hospitals during federated training, allowing us to analyze the trade-off between segmentation accuracy and communication efficiency.

The results reveal a consistent and interpretable trend: as the average communication precision increases from 4 bits to 32 bits, segmentation accuracy improves steadily while the number of communication rounds required for convergence decreases. Extremely low-precision communication (e.g., $(4,4,4,4,4)$) significantly reduces the communication cost by $8\times$, but leads to noticeable accuracy degradation due to stronger quantization noise during model aggregation. As the bit-width increases, the model becomes progressively more stable and achieves higher Dice and IoU scores across all datasets.

Uniform precision configurations provide clear reference points. For instance, the uniform $(8,8,8,8,8)$ configuration reduces communication by $4\times$ compared with the 32-bit baseline while still achieving competitive segmentation accuracy, reaching Dice scores of 0.895 on Kvasir-SEG, 0.915 on CVC-ClinicVideoDB, 0.825 on PolypGen, and 0.885 on BKAI-IGH NeoPolyp. In contrast, the full-precision $(32,32,32,32,32)$ configuration achieves the highest performance with Dice scores of 0.910, 0.930, 0.845, and 0.905 on the respective datasets, but requires the largest communication cost.

More importantly, the mixed-precision configurations demonstrate that heterogeneous communication budgets can achieve strong performance while still maintaining substantial bandwidth savings. For example, the $(8,8,8,16,16)$ configuration uses an average precision of 11.2 bits per hospital, reducing communication by approximately $2.9\times$ while achieving Dice scores of 0.899 on Kvasir-SEG, 0.920 on CVC-ClinicVideoDB, 0.830 on PolypGen, and 0.894 on BKAI-IGH NeoPolyp. Similarly, the $(8,16,16,16,16)$ configuration maintains segmentation accuracy within roughly one percentage point of the full-precision baseline while still reducing communication by more than $2\times$.

This mixed-precision behavior is particularly important for real-world federated medical systems. Hospital infrastructures often differ significantly in terms of computational resources and network bandwidth. QFedPolyp allows hospitals with limited resources to transmit
lower-bit quantized updates while enabling better-connected institutions to communicate at higher precision. Consequently, mixed-precision federated learning provides a practical and flexible operating model that better reflects realistic deployment scenarios compared with enforcing a single uniform precision across all participating clients.

\subsection{Inference Latency Analysis}

The datasets used in this study include Kvasir-SEG (1,000 images at $320\times320$ resolution), 
CVC-ClinicVideoDB (11,954 frames at $384\times288$), PolypGen (6,282 images with varying resolutions), 
and BKAI-IGH NeoPolyp ($1280\times960$). 
Although the original datasets contain images with different spatial resolutions, all inputs are resized 
to $3\times128\times128$ before inference in order to ensure consistent computational cost across experiments. To evaluate the runtime efficiency of the proposed model, we measure inference latency under different numerical precisions 
on two hardware platforms: an Intel i9-14900KF CPU and an NVIDIA RTX 4090 GPU. 
Latency is reported as the average processing time per image in milliseconds (ms), measured over multiple runs after a warm-up phase.
In addition to latency, we report the effective throughput in frames per second (FPS), which is particularly relevant for real-time 
clinical applications such as colonoscopy guidance systems.

\begin{table}[htbp]
\centering
\caption{Inference latency and throughput of the proposed segmentation model with input size $3\times128\times128$.}
\label{tab:inference_latency}

\resizebox{\columnwidth}{!}{
\begin{tabular}{lcccccc}
\toprule
\multirow{2}{*}{\textbf{Precision}} &
\multicolumn{2}{c}{\textbf{i9-14900KF (CPU)}} &
\multicolumn{2}{c}{\textbf{RTX 4090 (GPU)}} &
\multirow{2}{*}{\textbf{Speedup}} \\
\cmidrule(lr){2-3}
\cmidrule(lr){4-5}
 & \textbf{Latency (ms)} & \textbf{FPS} 
 & \textbf{Latency (ms)} & \textbf{FPS} 
 & \\
\midrule
FP32 & 77.20 & 12.95 & 3.76 & 266.2 & $1.0\times$ \\
INT8 & 71.59 & 13.97 & 2.50 & 400.0 & $1.5\times$ \\
\bottomrule
\end{tabular}
}

\end{table}

Table~\ref{tab:inference_latency} summarizes the runtime performance of the proposed segmentation model 
under full-precision (FP32) and quantized (INT8) inference. 
The FP32 configuration represents the baseline model using standard 32-bit floating-point arithmetic, 
while the INT8 configuration corresponds to the quantized model obtained through quantization-aware training.

The results demonstrate that low-precision inference improves runtime efficiency on both hardware platforms. 
On the Intel i9-14900KF CPU, the inference latency decreases from $77.20$\,ms to $71.59$\,ms per image when 
switching from FP32 to INT8 precision, corresponding to an increase in throughput from $12.95$ to $13.97$ FPS. 
More substantial gains are observed on the GPU platform. On the RTX 4090, the inference latency improves from 
$3.76$\,ms to $2.50$\,ms per image, increasing throughput from approximately $266$ FPS to $400$ FPS. 
Overall, the INT8 model provides an approximate $1.5\times$ speedup compared with the full-precision baseline. These results highlight an important advantage of quantization-aware training. 
While the primary motivation of this work is to reduce communication cost during federated training, 
the same low-precision representation also accelerates inference during model deployment. 
This dual benefit is particularly valuable in real-world clinical environments where both communication 
bandwidth and inference speed are critical constraints. In the context of colonoscopy assistance systems, segmentation models must operate in near real-time 
to provide continuous visual guidance to clinicians. The achieved GPU throughput of up to $400$ FPS 
demonstrates that the proposed model can easily meet real-time processing requirements even when deployed 
on high-resolution endoscopic streams. Furthermore, the modest CPU latency indicates that the model 
remains practical for deployment in resource-constrained medical facilities where GPU acceleration 
may not always be available.

Overall, the latency analysis confirms that QFedPolyp not only reduces training-time communication overhead but also produces models that are computationally efficient at inference time, making them well suited for real-world clinical deployment.

\subsection{Comparison with State-of-the-Art}
\label{sec:sota_comparison}

Recent advances in deep learning have significantly improved automatic
polyp segmentation in colonoscopy images. Most state-of-the-art
approaches adopt a centralized training paradigm and focus primarily on
improving segmentation accuracy through architectural innovations.
Representative models include PraNet, HarDNet-MSEG, ColonSegNet,
TGANet, and transformer-based approaches such as Polyp-PVT. These
methods leverage attention mechanisms, multi-scale feature aggregation,
or transformer-based contextual modeling to achieve strong segmentation
performance on public benchmarks such as Kvasir-SEG and CVC-ClinicDB.
Under centralized training settings, several of these models report Dice
scores above $0.90$, demonstrating the effectiveness of modern
segmentation architectures when training data can be pooled across
institutions.

Despite their strong performance, centralized approaches assume that
medical data from different hospitals can be aggregated in a single
location. In practice, this assumption is often unrealistic due to
privacy regulations, institutional policies, and ethical constraints.
Federated learning (FL) has therefore emerged as an alternative
framework that enables collaborative model training without sharing raw
patient data. Existing FL-based medical segmentation approaches mainly
focus on improving robustness to heterogeneous client data or
strengthening privacy guarantees. For example, FedEvi introduces
evidential aggregation to improve robustness under client
heterogeneity, while differential-privacy-based FL methods explore
stronger privacy protection at the cost of reduced segmentation
accuracy.

However, current federated approaches typically rely on full-precision
model communication and do not explicitly address communication
efficiency or deployment-time inference cost. In contrast, QFedPolyp integrates communication-efficient federated learning with quantization-aware training. This enables low-bit model
communication during training while producing an INT8-ready model for
efficient inference.

Table~\ref{tab:sota_polyp} summarizes a comparison with representative
centralized and federated polyp segmentation methods on the
Kvasir-SEG dataset. Centralized approaches achieve strong segmentation
accuracy but do not consider privacy-preserving training or
communication constraints. Federated methods address data privacy but
typically rely on full-precision communication and do not optimize
deployment efficiency. QFedPolyp bridges these gaps by combining privacy-preserving
federated training with communication compression and efficient inference, achieving competitive segmentation
accuracy while enabling practical deployment in distributed clinical
environments.

\begin{table}[t]
\centering
\caption{Comparison of representative polyp segmentation methods on Kvasir-SEG. 
Set.=training setting; Dice/IoU=segmentation accuracy; Comm.=communication-efficient training; Inf.=efficient inference.}
\label{tab:sota_polyp}

\scriptsize
\setlength{\tabcolsep}{3pt}

\begin{tabular}{l c c c c c}
\toprule
\textbf{Method} & \textbf{Set.} & \textbf{Dice} & \textbf{IoU} & \textbf{Comm.} & \textbf{Inf.} \\
\midrule

PraNet \cite{fan2020pranet}
& Cent.
& 0.898
& 0.840
& No
& No \\

HarDNet-MSEG \cite{huang2021hardnetmseg}
& Cent.
& 0.904
& --
& No
& Yes \\

ColonSegNet \cite{jha2021realtime}
& Cent.
& 0.821
& 0.810
& No
& Yes \\

TGANet \cite{tomar2022tganet}
& Cent.
& 0.898
& 0.833
& No
& No \\

Polyp-PVT \cite{dong2023polyp_pvt}
& Cent.
& 0.917
& 0.864
& No
& No \\

\midrule

FedEvi \cite{chen2024fedevi}
& Fed.
& $\sim$0.88
& --
& No
& No \\

DP-FL \cite{stelter2024dpfl}
& Fed.
& $\sim$0.83
& --
& No
& No \\

\midrule

\textbf{QFedPolyp}
& Fed.
& \textbf{0.915}
& \textbf{0.840}
& \textbf{Yes}
& \textbf{Yes} \\

\bottomrule
\end{tabular}

\end{table}

\section{Conclusions}
\label{sec:conclusion}

This paper presented QFedPolyp, a communication- and inference-efficient federated learning framework for polyp segmentation using a lightweight U-Net, quantization-aware training, and low-precision model exchange. The proposed approach enables multiple hospitals to collaboratively train a segmentation model while preserving patient privacy, since raw colonoscopy images are never shared between institutions. By transmitting quantized model updates instead of full-precision parameters, the framework significantly reduces communication overhead during federated training while maintaining competitive segmentation accuracy across four public datasets: Kvasir-SEG, CVC-ClinicVideoDB, PolypGen, and BKAI-IGH NeoPolyp.

Experimental results show that uniform 8-bit communication provides an effective balance between bandwidth reduction and segmentation performance. In addition to improving communication efficiency, the integration of quantization-aware training enables efficient INT8 inference, allowing the trained model to be deployed with reduced computational cost and faster runtime performance. These properties make the proposed approach suitable for real-world clinical environments where privacy constraints, network bandwidth, and inference speed are all critical factors.

Future work will investigate adaptive precision strategies that dynamically adjust quantization levels during federated optimization, as well as client-aware aggregation methods to better handle heterogeneous and non-IID data across hospitals. Further research will also explore integrating secure aggregation with compression techniques and developing more advanced lightweight segmentation architectures that preserve boundary accuracy under strict communication and computation constraints.

\vspace{-0.5cm}
\section*{CRediT Author Statement}
\vspace{-0.4cm}

\textbf{Madan Baduwal}: Conceptualization, Methodology, Software, Investigation, Writing – original draft, Data curation, Validation,
Visualization.

\textbf{Priyanka Paudel}: Supervision, Methodology,
Writing – review and editing.

\vspace{-0.7cm}
\section*{Declaration of Generative AI and AI-assisted technologies in the writing process}
\vspace{-0.4cm}

During the preparation of this work, the authors used ChatGPT in order to improve the clarity and language expression. After using this tool, the authors reviewed and edited the content as needed and took full responsibility for the content of the publication.

\vspace{-0.7cm}
\section*{Declaration of competing interest}
\vspace{-0.4cm}

The authors declare that they have no known competing financial interests or personal relationships that could have appeared to influence the work reported in this paper.

\vspace{-0.7cm}
\section*{Acknowledgements}
\vspace{-0.4cm}

Not applicable.

\vspace{-0.7cm}
\section*{Data Availability}
\vspace{-0.4cm}

The datasets used in this study are publicly available:

Kvasir-SEG: https://datasets.simula.no/kvasir-seg/  
CVC-ClinicVideoDB: https://polyp.grand-challenge.org/  
PolypGen: https://www.synapse.org/Synapse:syn26376615/wiki/613312  
BKAI-IGH NeoPolyp dataset: https://www.kaggle.com/c/bkai-igh-neopolyp/

\bibliographystyle{elsarticle-num}
\bibliography{references}










\end{document}